# An Automatic Seeded Region Growing for 2D Biomedical Image Segmentation

Mohammed. M. Abdelsamea

Mathematics Department, Assiut University, Egypt

**Abstract.** *In this paper, an automatic seeded region growing algorithm is proposed for cellular image segmentation. First, the regions of interest (ROIs) extracted from the preprocessed image. Second, the initial seeds are automatically selected based on ROIs extracted from the image. Third, the most reprehensive seeds are selected using a machine learning algorithm. Finally, the cellular image is segmented into regions where each region corresponds to a seed. The aim of the proposed is to automatically extract the Region of Interests (ROI) from the cellular images in terms of overcoming the explosion, under segmentation and over segmentation problems. Experimental results show that the proposed algorithm can improve the segmented image and the segmented results are less noisy as compared to some existing algorithms.*

## 1. Introduction

Quantitative studies have been performed based on populations of biological images. Such studies extremely require methods for segmentation, feature extraction, and classification. A first step in many analysis pipelines is segmentation, which can occur at several levels (e.g., separating nuclei, cells, tissues). This task has been an active field of research in image processing over the last 30 years, and various methods have been proposed and analysed depending on the modality, quality, and resolution of the microscopy images to analyze. For image segmentation, there are four main approaches: threshold techniques, boundary based techniques, region based techniques, and hybrid techniques [1].

In threshold techniques [2], all pixels whose value (grey level, colour value, or other) lie within a certain range can be classified as one class. Such methods neglect all of the spatial information of the image and do not cope well with.

Boundary based methods [3] use the postulate of the rapid changes of the pixel values at the boundary between two regions. The basic method here is to apply a gradient operator (e.g., Sobel, Roberts filter). High values of this filter provide candidates for region boundaries, which must then be modified to produce closed curves representing the boundaries between regions. Converting the edge pixel candidates to boundaries of the regions of interest is a difficult task. The complement of the boundary-based approach is to work with the regions.

Region based methods [4] rely on the postulate that all neighbouring pixels within the one region have similar value or a specific range. This leads to the class of algorithms known as region growing of which the "split and merge" technique is probably the best known. The general procedure is to compare a specific feature of one pixel to its neighbour(s) feature. If a criterion of homogeneity is satisfied, the pixel is classified to the same class as one or more of its neighbours. The choice of the homogeneity criterion is critical for even moderate success and in all instances the results are upset by noise. The fourth type is the hybrid techniques [5] which combine boundary and region criteria. This class includes morphological watershed segmentation and variable-order surface fitting. The watershed method is generally applied to the gradient of the image. This gradient image can be viewed as topography with boundaries between regions as ridges. Segmentation is equivalent to flooding the topography from the seed points with region boundaries being erected to keep water from different seed points from meeting. As same as all pervious method the technique encounters difficulties with images in which regions are both noisy and have blurred or indistinct boundaries. Furthermore, the method is also computationally very expensive. In this paper, an accurate segmentation method is proposed based on the seeded region growing method to be less sensitive to the noisy image and quantified and avoid explosion, leaking, under segmentation, and over segmentation problems.

## 2. Related Work

Jie Wu et al [6] proposed a new texture feature-based seeded region growing algorithm for automated segmentation of organs in abdominal MR images. The proposed seed point determination method is based on a cost-minimization approach. An ideal candidate seed point should have these properties: i) It should be inside the region and near the centre of the region ii) Assume most of the pixels in the ROI belong to the region, the feature of this seed point should be close to the region average iii) The distances from the seed pixel to its neighbours should be small enough to allow continuous growing.

Mehnert and Jackway [7] pointed out that Seeded Region Growing (SRG) has two inherent pixel order dependencies that cause different resulting segments. The first-order dependency occurs whenever several pixels have the same difference measure to their neighbouring regions. The second-order dependency occurs when one pixel has the same difference measure to several regions. They used parallel processing and re-examination to eliminate the order dependencies.

Frank et al [8] present an automatic seeded region growing algorithm for color image segmentation. In this paper, they apply the SRG to color images with automatic seed selection. they develop strategies to avoid the two order dependencies. For automatic seed selection, the following three criteria must be satisfied. First, the seed pixel must have high similarity to its neighbors. Second, for an expected region, at least one seed must be generated in order to produce this region. Third, seeds for different regions must be disconnected.

## 3. Methodology Overview

The system consist of three stages: i) After the preprocessed image is loaded into the system, the ROIs extracted via Otsu image segmentation from the image as a binary object; ii) a proposed method is applied for automatically selected seeds based on positions and intensities of pixels located in ROIs extracted; iii) a region growing procedure is recursively called for segment an input image as a seeded region growing based on seeded candidates and Otsu threshold. Fig 1 illustrates the overview of the algorithm.

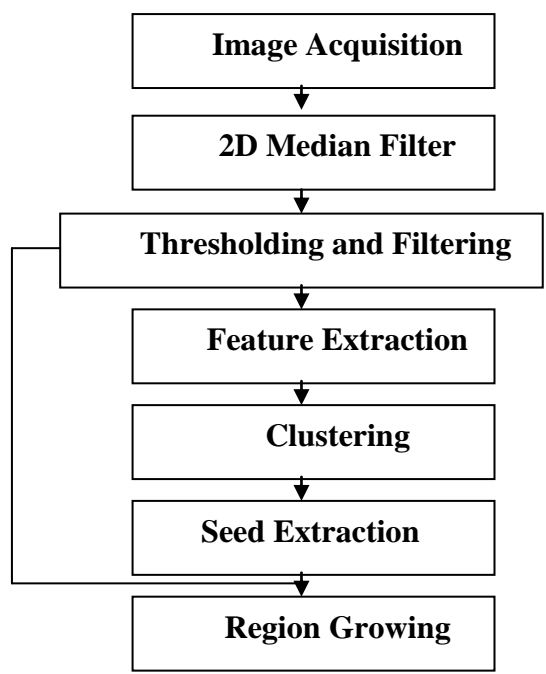

Fig. 1. Outline of the proposed algorithm.

## The Method for Automatic Seed Selection

For automatic seed selection, an ideal candidate must have the following properties and satisfy the following criteria.

i) Make sure that the seed point candidate is not on the boundary of two regions and/or not outlier.

ii) Check whether this candidate seed point has high similarity to its neighbours, which it should be inside the selected ROI and has the highest similarity between its neighbours to accurately represent the region.

iii) At least one seed pixel must be generated for each extracted ROI.

iv) The distances measured from the seed pixel to its neighbours should be inside extracted region to allow continuous growing.

For satisfy these criteria the proposed segmentation describes as follow:

The first segmentation step after image acquisition tries to reduce the noise present in the image. To reduce noise several non-linear filters can be employed. One of the simplest techniques, the median filter, can be employed here given that it provided good noise reduction without affecting the borders of the objects on the image.

Second, an Otsu segmentation method is applied in order to: extract ROIs from the input image and extract Otsu threshold, threshold that classifies the image into two clusters such that we minimize the area under the histogram for one cluster that lies on the other cluster's side of the threshold, to be used later as input in region growing procedure. The aim of the filtering step is to remove outliers points and points located on the boundary of ROIs by considering and using R X R neighbourhood mask to make sure that all the tested seed point's neighbours located within this mask, A particular pixel will be considered as a boundary pixel or outlier if at least one of all its neighbours is determined as a background pixel. The size of the neighbourhood to be considered around the pixel is determined by a radius value R.

Third, candidate seeds positions and its intensity values are extracted as features to represent the training set, which in this case the candidate seeds are the patterns of the training set, and apply it for clustering via machine learning technique; in this work K-mean clustering technique has been used.

After training process, the training pattern is applied as a test pattern to the learned algorithm and K is specified accordingly, which K is number of seeds or number of object that user would like to extract, to get the most reprehensive point to use as seeds for the Region Growing method for segment image into region where each region corresponds to a seed.

In Seed Extraction, we can consider extracting of the seed as a Missing Value Imputation problem which can be solved by extracting the prototypes of the positions of the centroids to act as an ideal seeds. By this step we make sure that this candidate seed has the highest similarity between its neighbours, at least one seed must be generated for each expected region, and the distances measured from the seed pixel to its neighbours should be inside extracted region to allow continuous growing.

It's obviously clear that this proposed method can avoid the explosion, under-segmentation and over-segmentation problems.

How to quantify an explosion?

By using the concept of considering the neighbours for growing the region, the object will not grow over its boundary.

How to avoid under-segmentation?

1. Using Otsu threshold with the main concepts of the growing region strategy have been used in this work, allow extract the whole region and make a stop before explosion.

2. make sure that the seed point candidate is not on the boundary of two regions.

How to avoid over-segmentation?

At least one seed pixel must be generated for each expected region, and Make sure that this candidate seed has the highest similarity between its neighbours to accurately represent the region.

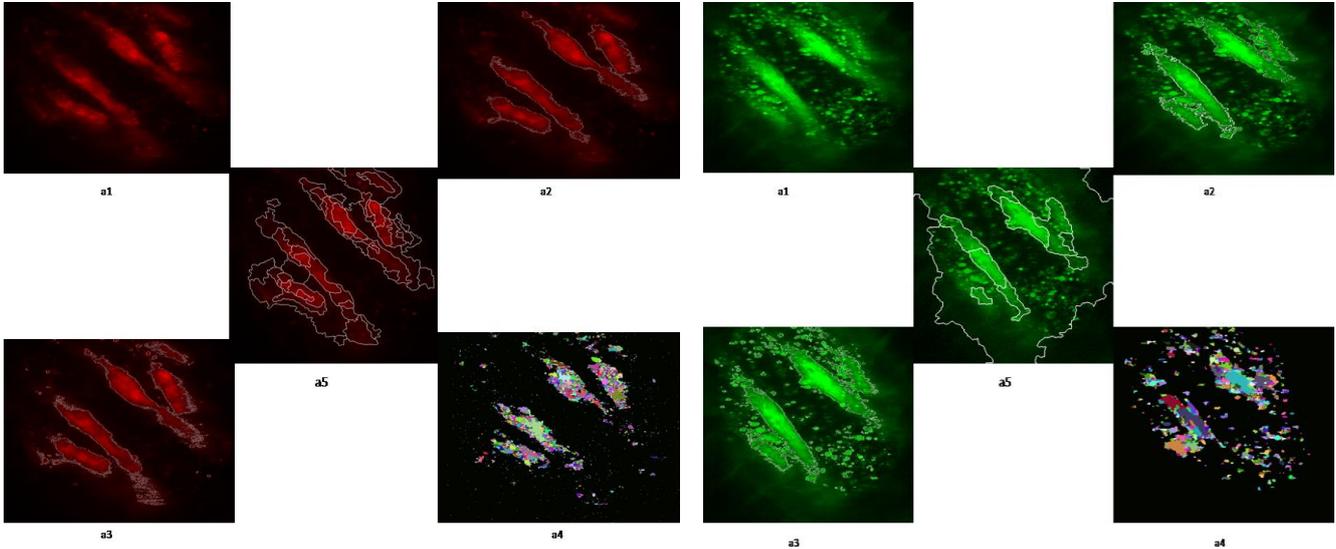

Fig. 2. (a1) Original image, (a2) the result of segmentation using the proposed algorithm with K = 4 and R = 3, (a3) the segmentation result using Otsu method, (a4) the segmentation result using watershed method with L = .05 and T = .05, (a5) the segmentation result using JSEG.

Fig. 3. (a1) Original image, (a2) the result of segmentation using the proposed algorithm with K = 4 and R = 3, (a3) the segmentation result using Otsu method, (a4) the segmentation result using watershed method with L = .08 and T = .08, (a5) the segmentation result using JSEG.

filtering which has been used in this work to avoid the explosion and under segmentation that might be happen because of the leaking problem of the region growing. For this reason a neighbourhood condition has been applied for accepting the pixel to be including in the region, these conditions based on the intensity value by considering an Otsu Threshold as an optimal threshold. This process is recursively executed until no more seeds can be tested. The second reason for considering the neighbourhood intensities instead of only the current pixel intensity is that small structures are less likely to be accepted in the region. The size of the neighbourhood to be considered around each pixel is defined by a user-provided integer radius.

## 4. Experimental Results

I have performed the experiments using the proposed segmentation algorithm and performed comparisons with some existing segmentation algorithm. In Fig 2, 3, and 4, we show the comparison with Otsu algorithm, Watershed algorithm, and JSEG algorithm.

By observing Fig 2(a3, a4, a5), Fig 3(a3, a4, a5), and Fig 4(a3, a4, a5), we can realize that their results are over segmented, and segmented objects in Fig 2(a5), Fig 3(a5), and Fig 4(a5) are inappropriately merged to background.

By comparing these results with results of our proposed algorithm in Fig 2(a2), Fig 3(a2), and Fig 4(a2) we can clearly notice that the proposed segmentation algorithm overcoming the over-segmentation problem and is less noisy as compared to other algorithms.

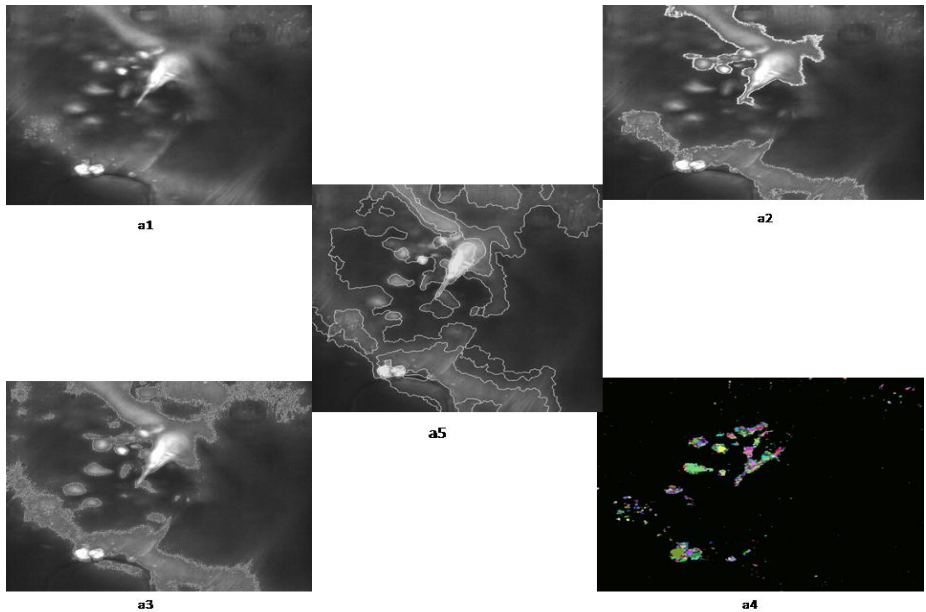

Fig. 4. (a1) Original image, (a2) the result of segmentation using the proposed algorithm with K = 3 and R = 3, (a3) the segmentation result using Otsu method, (a4) the segmentation result using watershed method with L = .5 and T = .5, (a5) the segmentation result using JSEG.

## 5. Conclusions

In this paper, an automatic seeded region growing framework for automatically segment 2D cellular image is proposed. One of the advantages of this algorithm is obvious, as it provides a parameter free production environment to allow minimum user intervention. This can be especially helpful for batch work or to novice computer users. The other benefit of this algorithm is overcoming the explosion, under segmentation and over segmentation problems. Furthermore, it is clearly that the proposed algorithm able to consuming time because only specific regions are used to produce seeds. However, the proposed algorithm does have drawbacks. Only pixel intensity and its position are extracted as features and were used as criteria to selecting seeds and growing the region